%% file: main_eccv.tex
\documentclass[runningheads]{llncs}
\usepackage{graphicx}

\usepackage{wrapfig}

\usepackage{tikz}
\usepackage{comment}
\usepackage{amsmath,amssymb} 
\usepackage{color}

\begin{document}
\pagestyle{headings}
\mainmatter
\def\ECCVSubNumber{7}  

\title{Properties Of Winning Tickets On Skin Lesion Classification} 

\titlerunning{Properties of Winning Tickets On Skin Lesion Classification}
%
\author{Sherin Muckatira}
\authorrunning{S. Muckatira}
%
\institute{smuckati@asu.edu}
\maketitle

\begin{abstract}
Skin cancer affects a large population every year - automated skin cancer detection algorithms can thus greatly help clinicians. Prior efforts involving deep learning models have high detection accuracy. However, most of the models have a large number of parameters, with some works even using an ensemble of models to achieve good accuracy. In this paper, we investigate a recently proposed pruning technique called Lottery Ticket Hypothesis. We find that iterative pruning of the network resulted in improved accuracy, compared to that of the unpruned network, implying that - the lottery ticket hypothesis can be applied to the problem of skin cancer detection and this hypothesis can result in a smaller network for inference. We also examine the accuracy across sub-groups - created by gender and age - and it was found that some sub-groups show a larger increase in accuracy than others. 
\end{abstract}

\section{Introduction}
\label{submission}

Pattern recognition is used in many fields of medicine including dermatology: dermatologists look for specific patterns in the skin such as asymmetry, irregular border, color, diameter and evidence of growth, in addition to patients overall skin and medical history, to classify skin lesions as cancerous \cite{topol2019deep}.
Around 5 million skin cancer cases occur annually in the United States \cite{SkinCancer}; early screening and detection of cancer can significantly reduce the mortality rate \cite{armstrong1987malignant} and in most cases result in 100\% recovery. Hence, automatic classification of skin lesions can greatly benefit the medical  community by offloading some of their diagnostic efforts. \\
International Skin Imaging Collaboration (ISIC) has organized challenges, for the past few years, to improve the accuracy of skin disease diagnosis; it has proposed standards for creating skin lesion images and provides a dataset comprising of skin lesion images from different clinical centers. The images in the 2019 ISIC challenge\cite{codella2018skin,tschandl2018ham10000,combalia2019bcn20000} can fall into one of the following eight disease categories: melanoma, melanocytic nevus, basal cell carcinoma, actinic keratosis, benign keratosis, dermatofibroma, vascular lesion, and squamous cell carcinoma. The goal of the ISIC challenge is to improve the accuracy of skin cancer prediction. A literature survey of state-of-the-art solving this particular challenge \cite{gessert2020skin,pacheco2019skin} show that a combination of multiple models (ensemble learning) provides better accuracy results. Since the dataset has only few thousands of images, most of the approaches use transfer learning to overcome this limitation. The future of AI in skin lesion analysis would allow patients to take images of skin lesions using a smartphone  \cite{topol2019deep}; the AI algorithm would then be able to provide a diagnosis - aiding the dermatologist in focusing on efficient treatment. To use smartphones for diagnosing skin cancer, we would need to focus on the efficacy of storage and size of the network; this paper shows that pruning based on Lottery ticket hypothesis and transfer learning can be used to create a model which has fewer parameters and thus computationally efficient. Surprisingly, we observe that pruning results in improved accuracy.

\section{Method}
Traditional machine learning algorithms use hand-engineered features; the performance of these models highly depend on the quality of the features chosen. Deep learning, on the other hand extracts features from the input data. Deep Learning models need to be exposed to a large amount of data to be able learn all the representations in underlying data - this becomes an issue when such a large dataset is not available. Input data collected for medical tasks - which are usually from consenting patients who may have a certain condition - are expensive. Transfer learning - a technique where models trained on one task are used for a similar task - is used to overcome the challenge of working with a limited amount of data \cite{yosinski2014transferable}. The model is trained in a domain, source domain, where a large number of samples are available and this training aids the model in learning some of the generic features in the training data. The model can then transfer its knowledge to another domain, target domain, where very few samples are available. In the target domain the model can be used without having to train it from scratch: eliminating the requirement for having a very large dataset in this domain \cite{tan2018survey}.  The initial layers of the model extract the generic features in images, the last few layers learn classification specific to the task at hand - hence, while using transfer learning the last few layers need to be adjusted to suit the specific classification task and these layers need to be re-trained from scratch. Since the number of unique skin lesion images are too small to train a deep neural network, we leverage resnet-18: a pre-trained model trained on ImageNet. \cite{frankle2018lottery} shows that a dense network can be pruned by about 10\% - 20\% and the resulting network after pruning can: achieve training/test accuracy  better than the original network and train as quickly as the dense network; to mitigate the lower training accuracy of smaller networks, the sub-network surviving the pruning should retain the initial weights from the original dense network. Thus hypothesizing that SGD trains a sub-network (winning tickets) that is initialized well i.e. have won the initialization lottery. Pruning reduces the parameter count and thereby improves the energy consumption and storage requirements of the network - it is done by removing the lowest magnitude weights. In this paper we prune the network and train it on the ISIC 2019 dataset to obtain a sparse network - the results show that the test accuracy improves with pruning.
Pruning is performed as follows:
    \begin{enumerate}
        \item Initialize the dense network randomly.
        \item Train the dense network for a certain number of epochs.
        \item Prune p\% of the weights which are below a certain threshold.
        \item Re-initialize the surviving network with the initial weights from the dense network and repeat steps 2 through 4.
    \end{enumerate}

Prior works such as \cite{mahbod2020investigating} show that for datasets containing less number of examples, resnet18 performs better than deeper models such as resnet50 or densenet121; hence, we use a resnet18 model\cite{he2016deep} pre-trained on ImageNet as the backbone network.  The last fully connected layer of resnet18 is removed and two fully connected layers with 256 and 8 units (the number of classes in ISIC dataset) are added. We use Relu non-linearity and Dropout layer with dropout rate = 0.4. For the first stage of training without any pruning (L0), we freeze all the layers of the resnet except the last resnet block and the fully connected layers added at the end of the resnet block. For all subsequent pruning rounds, L1-L9, we unfreeze all the parameters of resnet blocks and let the pruning take place across all parameters in the model including the resnet backbone. We decide to prune by 18\% based on the experiments in the paper on Lottery Ticket Hypothesis \cite{frankle2018lottery}. The network is iteratively pruned by 2 percent in each level for 10 rounds (L0-L9); in each round the model is trained for 20 epochs. Pruning is done by zeroing weights below a particular threshold globally. We use Adam optimizer \cite{kingma2014adam} with a learning rate of 0.001 and weight decay of 1e-5.

The 2019 ISIC dataset comes with Metadata, age and gender, which aids in creation of subgroups - male, female, ages1-30, ages31-60, and ages 61-90. We see that pruning - based on Lottery ticket hypothesis -  is more biased in some groups; the greatest improvements in accuracy seen in the female and ages61-90 subgroups.

\begin{figure}
    \centering
    \includegraphics[height=4cm]{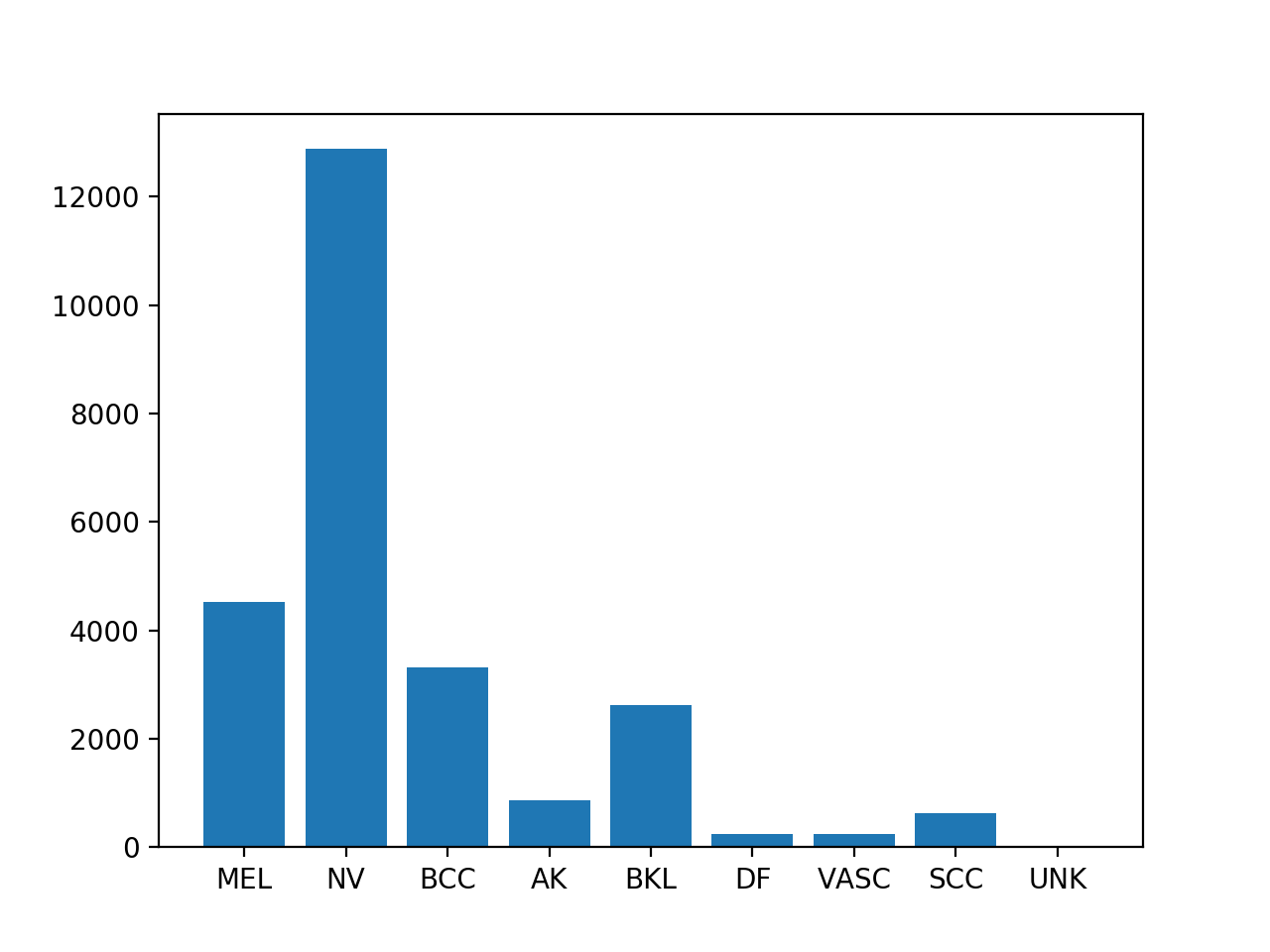}
    \caption{Number of samples in different classes.}
\end{figure}

\section{Experiments}
\subsection{Dataset and Pre-processing}
The dataset provided by the ISIC 2019 challenge is divided into different sub-groups based on age and gender. Imbalance in the dataset, 70\% of examples indicating class melanocytes nevus, is overcome by oversampling from classes which have lesser number of samples during training - ensuring that the model sees a better balance among training samples. There are 25000 images in the dataset, but the resnet18 model is fairly complex; to avoid overfitting we perform data augmentation using random horizontal flip - flipping a lesion horizontally creates a new image and the CNNs in the resnet are invariant to such flips. The image distribution among the different classes are shown in the figure 1, which shows that most of the images belong to the class melanocytic nevus(NV) and melanoma(MEL) is the next dominant class; classes DF and VASC have very few images. If the dataset is used as is, the model would predict the dominant class, melanocytic nevus, all the time - as it would have seen most examples from the dominant class. We use a weighted sampler which presents equal samples from all classes in each training batch. All the images are cropped to size (224,224) to be fed into resnet18 and normalized.

\begin{table*}
\centering
\scalebox{1.25}{
\begin{tabular}{c|c|c|c|c|c|c|c|c|c|c}
\hline
Subgroups & L0 & L1 & L2 & L3 & L4 & L5 & L6 & L7 & L8 & L9\\
\hline
\hline
Male & 54.49 & 60.23 & 60.42 & 59.17 & 60.54 & 60.91 & 61.97 & 61.59 & 62.72 & 62.12 \\
Female & 56.08 & 62.10 & 63.21 & 61.89 & 62.91 & 62.78 & 65.47 & 65.47 & 64.40 & 66.02 \\
Ages 1-30 & 66.67 & 63.5 & 68.0 & 60.67 & 68.17 & 64.83 & 66.17 & 66 & 67.33 & 70.83 \\
Ages 31-60 & 61.24 & 65.5 & 66.69 & 65.61 & 67.03 & 64.98 & 68.71 & 68.00 & 68.37 & 68.48 \\
Ages 61-90 & 41.71 & 53.41 & 51.65 & 53.59 & 50.88 & 55.82 & 54.83 & 55.35 & 54.53 & 54.35 \\
\hline
\end{tabular}}
\caption{Accuracy across sub-groups.}
\end{table*}

\subsection{Results}
The paper compares the accuracy of the subgroups, based on age and gender,  before and after pruning. L0 shows the accuracy without any pruning; L1-L9 show the accuracies across different levels of iterative pruning - at each stage the lowest 2\% weights which lie below a threshold are pruned. The affects of pruning on different sub-groups are different - subgroups female and ages 61-90 record the highest improvements. The male and the female sub-groups initially at L0 have accuracy spaced apart by 2\% but after iteratively pruning the network, the difference between accuracy of these two sub-groups spread apart by 4\% - indicating that pruning doesn't act on differnt sub-groups in the same way. In some sub-groups pruning reduces the difference between accuracies, e.x., sub-groups ages 1-30 and ages 61-90 have an initial difference in accuracy of the unpruned network by 25\% but after iterative pruning, this gap gets closer and the difference in accuracies is 16\%. Thus we can conclude that pruning has different affects on different sub-groups. Figure 2 shows the confusion matrix for all disease classes of the pruned network - it can be concluded by looking at the diagonal elements, true positives of a class, that the prediction of each class is predicted with highest accuracy that it belongs to that class; the false positives of each disease class is small compared to the true positive of that class.  It can be seen that Squamous Cell Carcinoma and Basal Cell Carcinoma have high recall. Classes Vascular and Melanoma have lowest recall; with class Vascular being mostly confused as Benign Keratosis and Melanoma being mostly confused as Actinic Keratosis. Most of the classes with low recall are being confused by Benign Keratosis class.

\begin{figure}
    \centering
    \includegraphics[height=6cm]{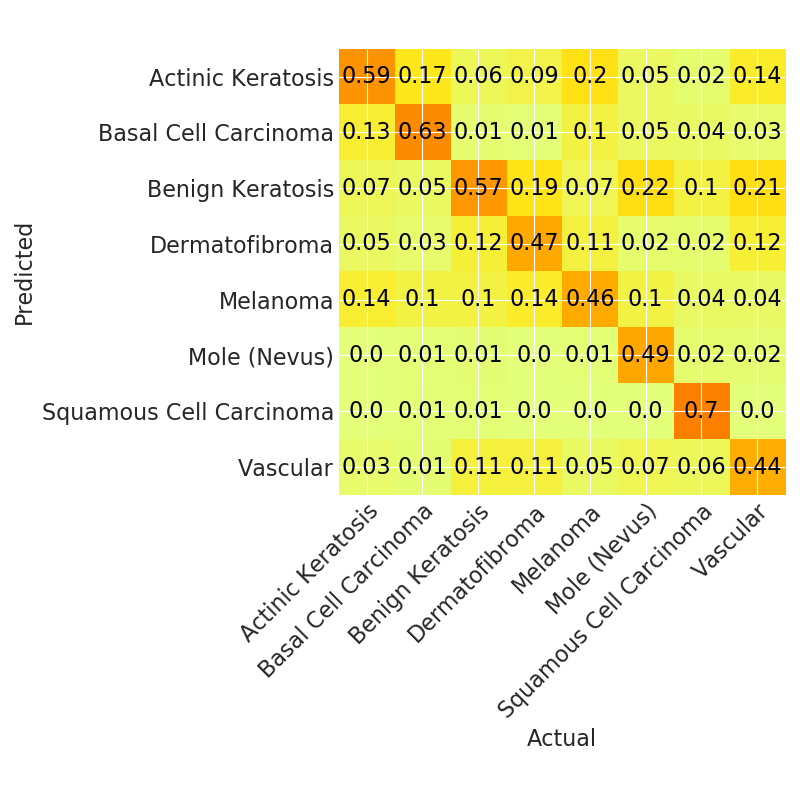}
    \caption{Confusion matrix of all disease classes. }
\end{figure}

\section{Conclusion}
In this paper, we compare the test accuracy of the unpruned network with various stages of pruning among different sub-groups - it was found that the test accuracy of the pruned network is better than that of the unpruned network. As part of future work we would like to improve the accuracy of the baseline network; some of the existing works show that improving the color constancy of the different images result in better accuracy. Since most of the existing literature use ensemble models to get better accuracy, we would like to use an ensemble model as the baseline  model and then prune it by 50-90\% to get better accuracy results. We have used resnet18 as the backbone of the model, it would be interesting to compare how the findings would differ if a different base model is used for transfer learning.


\bibliographystyle{splncs04}
\bibliography{main_eccv}

\newpage
\appendix

\input{appendix.tex}



\end{document}

%% file: appendix.tex
\section*{Appendix}

\textbf{Evolution of the Confusion Matrix across pruning levels}
\begin{figure}
    \centering
    \includegraphics[height=6cm]{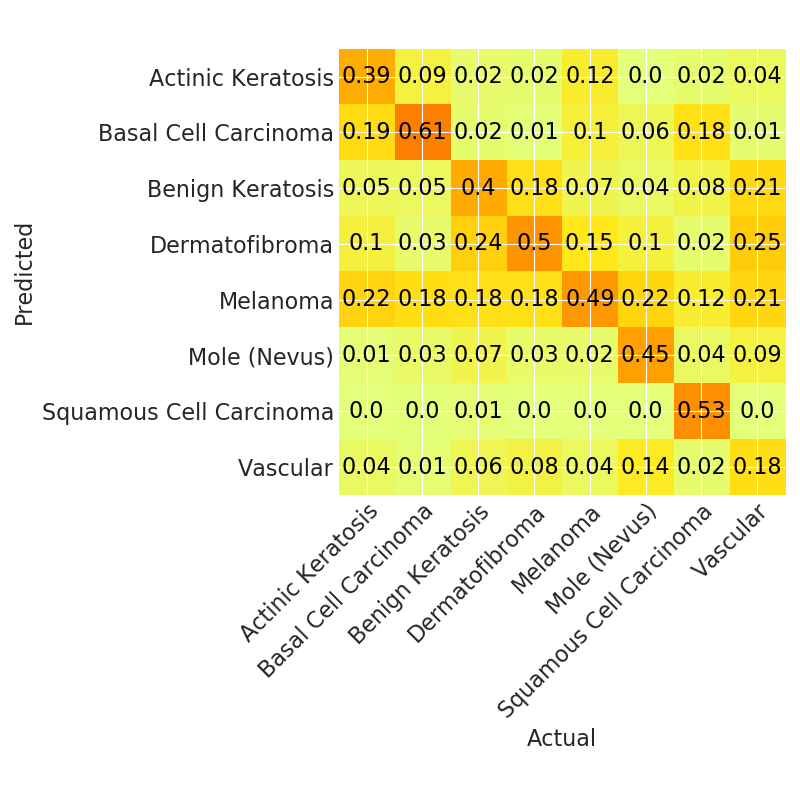}
    \caption{Confusion matrix of the unpruned network. }
\end{figure}

\begin{figure}
    \centering
    \includegraphics[height=6cm]{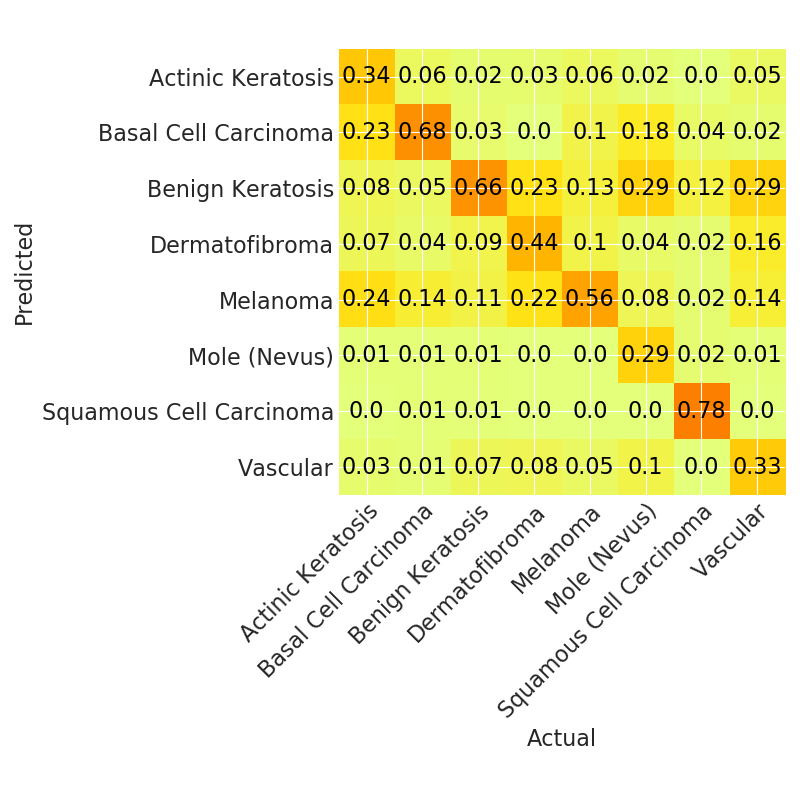}
    \caption{Confusion matrix of the network with 2\% pruning. }
\end{figure}

\begin{figure}
    \centering
    \includegraphics[height=6cm]{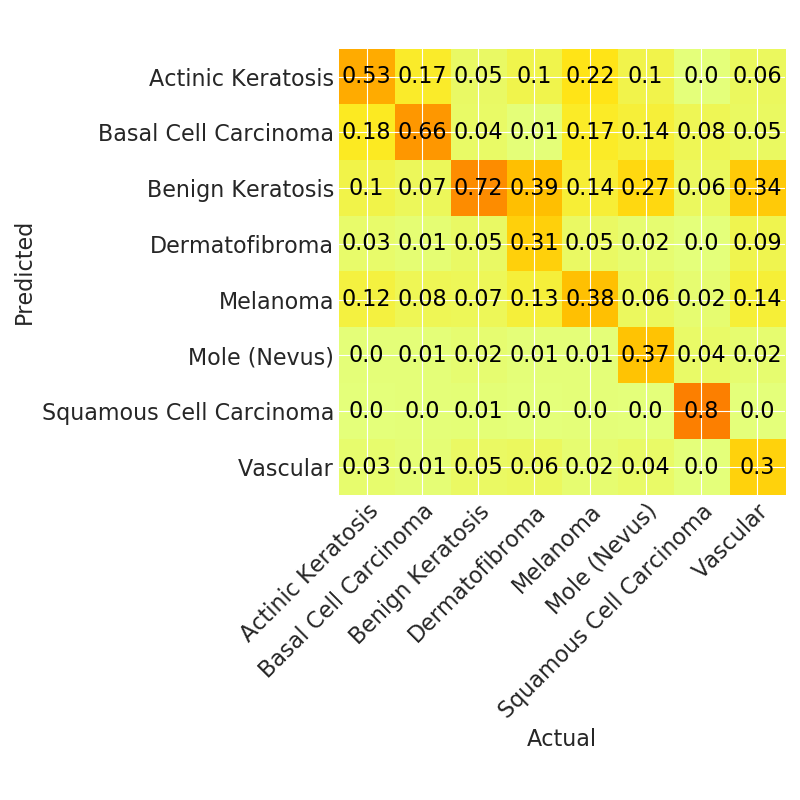}
    \caption{Confusion matrix of the network with 4\% pruning. }
\end{figure}

\begin{figure}
    \centering
    \includegraphics[height=6cm]{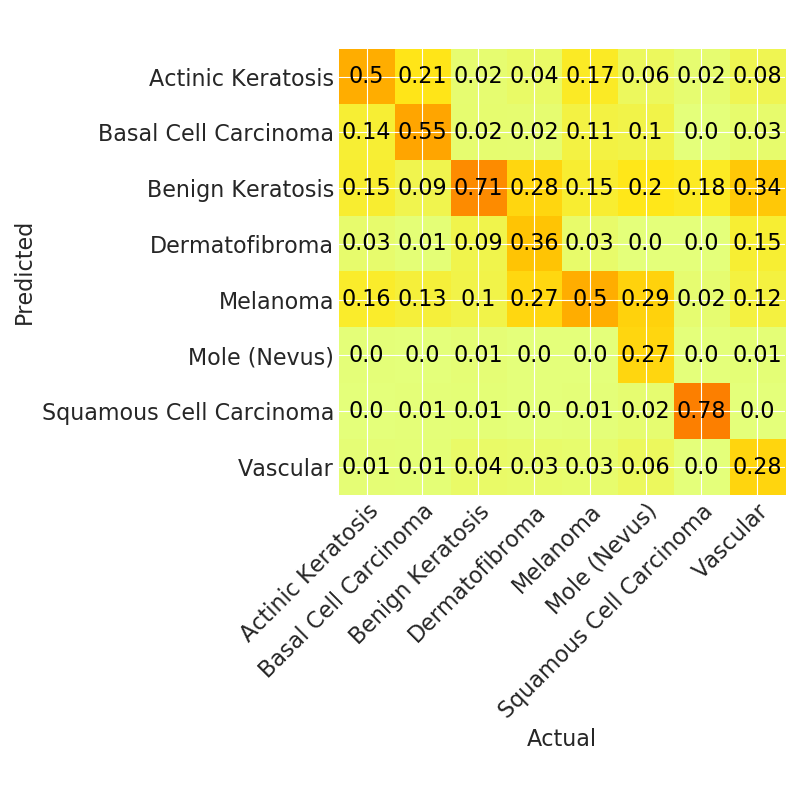}
    \caption{Confusion matrix of the network with 6\% pruning. }
\end{figure}

\begin{figure}
    \centering
    \includegraphics[height=6cm]{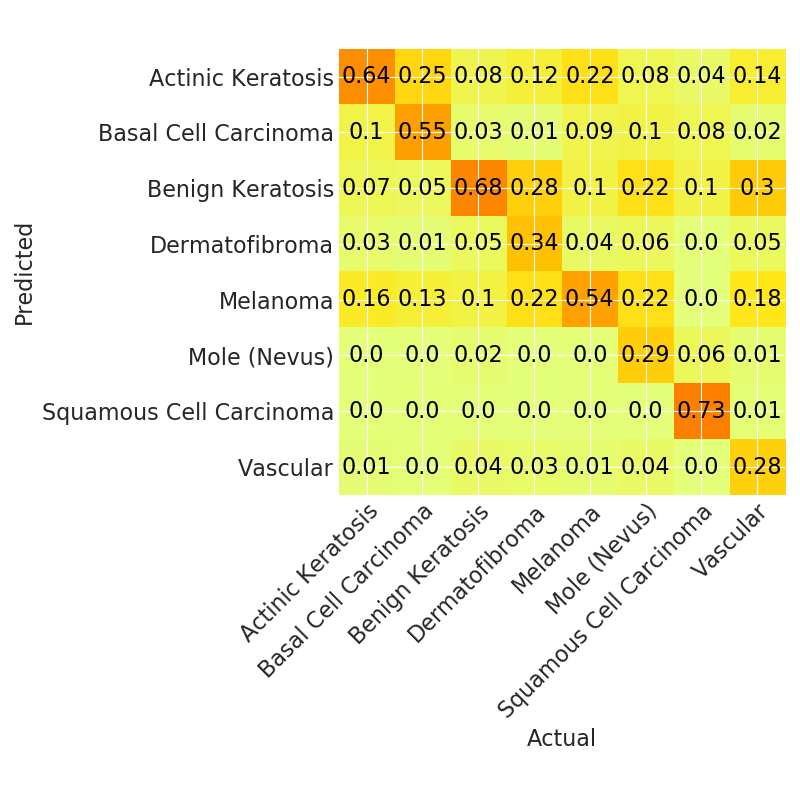}
    \caption{Confusion matrix of the network with 8\% pruning. }
\end{figure}

\begin{figure}
    \centering
    \includegraphics[height=6cm]{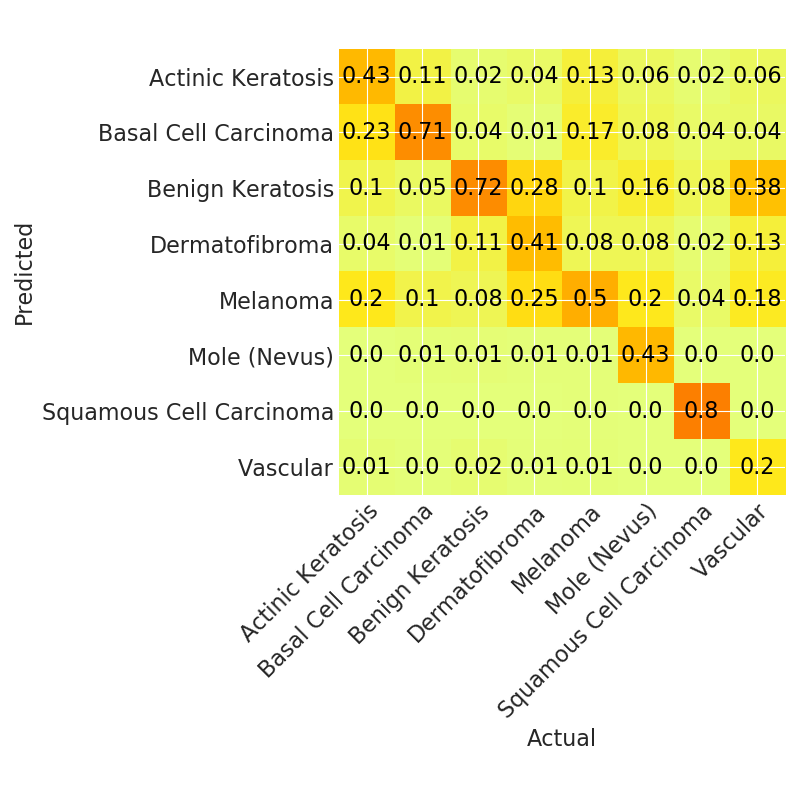}
    \caption{Confusion matrix of the network with 10\% pruning. }
\end{figure}

\begin{figure}
    \centering
    \includegraphics[height=6cm]{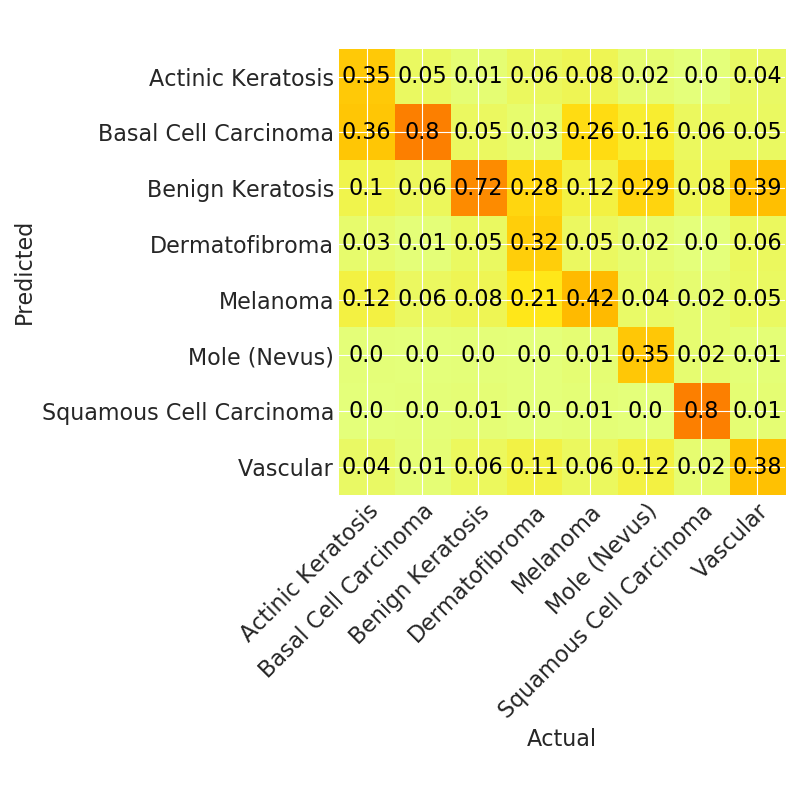}
    \caption{Confusion matrix of the network with 12\% pruning. }
\end{figure}

\begin{figure}
    \centering
    \includegraphics[height=6cm]{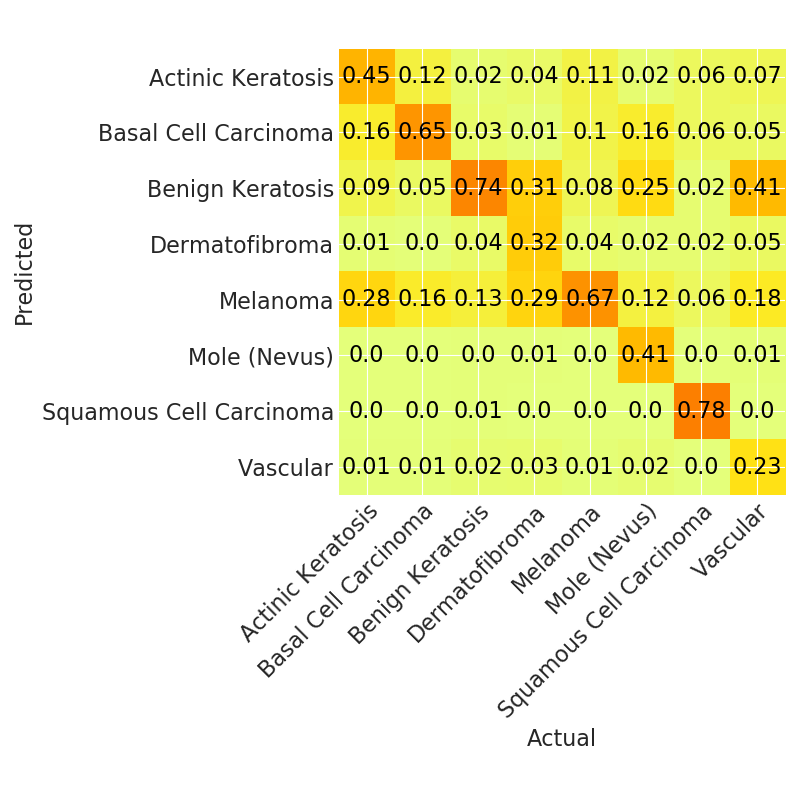}
    \caption{Confusion matrix of the network with 14\% pruning. }
\end{figure}

\begin{figure}
    \centering
    \includegraphics[height=6cm]{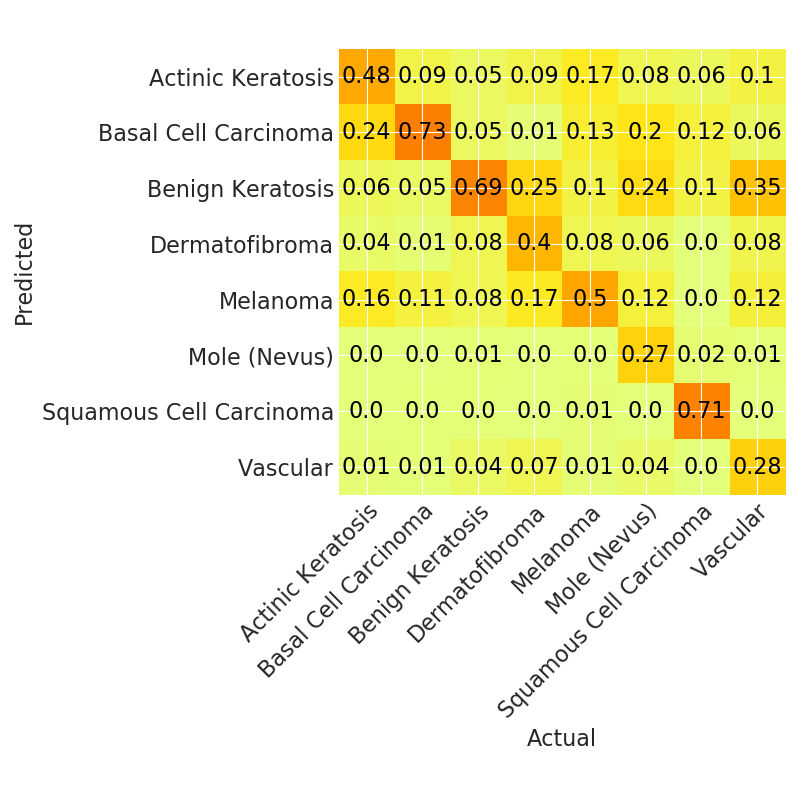}
    \caption{Confusion matrix of the network with 16\% pruning. }
\end{figure}

\begin{table*}
\centering
\scalebox{1.25}{
\begin{tabular}{c|c|c|c|c|c|c|c|c|c|c}
\hline
Classes & L0 & L1 & L2 & L3 & L4 & L5 & L6 & L7 & L8 & L9\\
\hline
\hline
Actinic Keratosis & 39 & 34 & 53 & 50 & 64 & 43 & 35 & 45 & 48 & 59 \\
Basal Cell Carcinoma & 61 & 68 & 66 & 55 & 55 & 71 & 80 & 65 & 73 & 63 \\
Benign Keratosis & 40 & 66 & 72 & 71 & 68 & 72 & 72 & 74 & 69 & 57 \\
Dermatofibroma  & 50 & 44 & 31 & 36 & 34 & 41 & 32 & 32 & 40 & 47 \\
Melanoma  & 49 & 56 & 38 & 50 & 54 & 50 & 42 & 67 & 50 &46 \\
Mole (Nevus)  & 45 & 29 & 37 & 27 & 29 & 43 & 35 & 41 & 27 & 49 \\
Squamous Cell Carinoma  & 53 & 78 & 80 & 78 & 73 & 80 & 80 & 78 & 71 & 70 \\
Vascular & 18 & 33 & 30 & 28 & 28 & 20 & 38 & 23 & 28 & 44 \\

\hline
\end{tabular}}
\caption{Evolution of True positives in each class across pruning levels}
\end{table*}

%% file: main_eccv.bbl
\begin{thebibliography}{10}
\providecommand{\url}[1]{\texttt{#1}}
\providecommand{\urlprefix}{URL }
\providecommand{\doi}[1]{https://doi.org/#1}

\bibitem{SkinCancer}
AmericanCancerSociety: Cancer facts and figures \url{https://bit.ly/2Lbu7Cz}

\bibitem{armstrong1987malignant}
Armstrong, B., Holman, C.: Malignant melanoma of the skin. Bulletin of the
  World Health Organization  \textbf{65}(2), ~245 (1987)

\bibitem{codella2018skin}
Codella, N.C., Gutman, D., Celebi, M.E., Helba, B., Marchetti, M.A., Dusza,
  S.W., Kalloo, A., Liopyris, K., Mishra, N., Kittler, H., et~al.: Skin lesion
  analysis toward melanoma detection: A challenge at the 2017 international
  symposium on biomedical imaging (isbi), hosted by the international skin
  imaging collaboration (isic). In: 2018 IEEE 15th International Symposium on
  Biomedical Imaging (ISBI 2018). pp. 168--172. IEEE (2018)

\bibitem{combalia2019bcn20000}
Combalia, M., Codella, N.C., Rotemberg, V., Helba, B., Vilaplana, V., Reiter,
  O., Halpern, A.C., Puig, S., Malvehy, J.: Bcn20000: Dermoscopic lesions in
  the wild. arXiv preprint arXiv:1908.02288  (2019)

\bibitem{frankle2018lottery}
Frankle, J., Carbin, M.: The lottery ticket hypothesis: Finding sparse,
  trainable neural networks. arXiv preprint arXiv:1803.03635  (2018)

\bibitem{gessert2020skin}
Gessert, N., Nielsen, M., Shaikh, M., Werner, R., Schlaefer, A.: Skin lesion
  classification using ensembles of multi-resolution efficientnets with meta
  data. MethodsX p. 100864 (2020)

\bibitem{he2016deep}
He, K., Zhang, X., Ren, S., Sun, J.: Deep residual learning for image
  recognition. In: Proceedings of the IEEE conference on computer vision and
  pattern recognition. pp. 770--778 (2016)

\bibitem{kingma2014adam}
Kingma, D.P., Ba, J.: Adam: A method for stochastic optimization. arXiv
  preprint arXiv:1412.6980  (2014)

\bibitem{mahbod2020investigating}
Mahbod, A., Schaefer, G., Wang, C., Ecker, R., Dorffner, G., Ellinger, I.:
  Investigating and exploiting image resolution for transfer learning-based
  skin lesion classification. arXiv preprint arXiv:2006.14715  (2020)

\bibitem{pacheco2019skin}
Pacheco, A.G., Ali, A.R., Trappenberg, T.: Skin cancer detection based on deep
  learning and entropy to detect outlier samples. arXiv preprint
  arXiv:1909.04525  (2019)

\bibitem{tan2018survey}
Tan, C., Sun, F., Kong, T., Zhang, W., Yang, C., Liu, C.: A survey on deep
  transfer learning. In: International conference on artificial neural
  networks. pp. 270--279. Springer (2018)

\bibitem{topol2019deep}
Topol, E.: Deep medicine: how artificial intelligence can make healthcare human
  again. Hachette UK (2019)

\bibitem{tschandl2018ham10000}
Tschandl, P., Rosendahl, C., Kittler, H.: The ham10000 dataset, a large
  collection of multi-source dermatoscopic images of common pigmented skin
  lesions. Scientific data  \textbf{5},  180161 (2018)

\bibitem{yosinski2014transferable}
Yosinski, J., Clune, J., Bengio, Y., Lipson, H.: How transferable are features
  in deep neural networks? In: Advances in neural information processing
  systems. pp. 3320--3328 (2014)

\end{thebibliography}
